# Parking Availability Prediction via Fusing Multi-Source Data with A Self-Supervised Learning Enhanced Spatio-Temporal Inverted Transformer

Yin Huang [a,#], Yongqi Dong [b,#], Youhua Tang [a], Li Li [c]

[a] *School of Transportation and Logistics, Southwest Jiaotong University, Chengdu, 610032, China*
Email: *huangyin@my.swjtu.edu.cn*; *tyhctt@swjtu.cn*

[b] *Institute of Highway Engineering, RWTH Aachen University, Aachen, 52074, Germany*
Email: *yongqi.dong@rwth-aachen.de*

[c] *Department of Automation, BNRist, Tsinghua University, Beijing, 100084, China*
Email: *li-li@tsinghua.edu.cn*

[#] These authors contributed equally to this work and should be considered as co-first authors.

**ABSTRACT**

The rapid growth of private car ownership has worsened the urban parking predicament, underscoring the need for accurate and effective parking availability prediction to support urban planning and management. To address key limitations in modeling spatio-temporal dependencies and exploiting multi-source data for parking availability prediction, this study proposes a novel approach with Self-supervised learning enhanced Spatio-Temporal Inverted Transformer (SST-iTransformer). The methodology leverages K-means clustering to establish parking cluster zones (PCZs), extracting and integrating traffic demand characteristics from various transportation modes (i.e., metro, bus, online ride-hailing, and taxi) associated with the targeted parking lots. Upgraded on vanilla iTransformer, SST-iTransformer integrates masking-reconstruction-based pretext tasks for self-supervised spatio-temporal representation learning, and features an innovative dual-branch attention mechanism: *Series Attention* captures long-term temporal dependencies via patching operations, while *Channel Attention* models cross-variate interactions through inverted dimensions. Extensive experiments using real-world data from Chengdu, China, demonstrate that SST-iTransformer outperforms baseline deep learning models (including Informer, Autoformer, Crossformer, and iTransformer), achieving state-of-the-art performance with the lowest mean squared error (MSE) and competitive mean absolute error (MAE). Comprehensive ablation studies quantitatively reveal the relative importance of different data sources: incorporating ride-hailing data provides the largest performance gains, followed by taxi, whereas fixed-route transit features (bus/metro) contribute marginally. Spatial correlation analysis further confirms that excluding historical data from correlated parking lots within PCZs leads to substantial performance degradation, underscoring the importance of modeling spatial dependencies. Overall, this work offers a robust solution for high-volatility parking prediction and provides actionable insights for developing data-driven intelligent urban parking management systems.

Keywords: Parking availability prediction, Deep learning, Clustering, Self-supervised learning, Spatio-Temporal Inverted Transformer, Multi-source data fusion

## 1. Introduction

### 1.1 Background

With rapid urbanization and the sustained increase in private vehicle ownership, parking scarcity has emerged as a critical concern in many metropolitan areas. Parking inaccessibility not only leads to longer driving times but also exacerbates urban traffic congestion, driver frustration, and environmental pollution (Bock et al., 2020; Jiang et al., 2025; Zeng et al., 2022). The challenge of finding parking spaces, especially during peak periods, is not solely due to the scarcity of parking spots but is also largely attributed to the lack of predictive analytics on parking availability (Reed et al., 2024; Y. Zhang et al., 2024). Parking dynamics, characterized by temporal fluctuations in arrivals and departures and the continual turnover of vehicles entering and exiting parking spaces, directly influence the availability of parking spots and introduce high volatility that complicates forecasting. Reliable, precise, and timely prediction of parking availability plays a crucial role in demand management, traveler information systems, and urban planning interventions (Choi and Lee, 2023), empowering city authorities and drivers to make better-informed decisions, optimizing the utilization of parking resources (Jiang and Fan, 2020; Zhang et al., 2023), facilitating parking slots sharing (Jiang et al., 2025; Xie et al., 2024), and thus helping alleviate traffic congestion.

To estimate and predict parking availability, some pioneering studies have been developed. Early approaches for parking availability prediction predominantly relied on probabilistic and statistical models. For example, Caliskan et al. (2007) proposed a continuous-time Markov chain-based model to predict future parking lot occupancy. While Rajabioun and Ioannou (2015) developed a spatio-temporal autoregressive framework leveraging real-time sensor data for forecasting parking availability. However, these sensor-based methodologies face significant limitations due to the inherent instability of parking detection systems (Perković et al., 2020), frequent equipment malfunctions (communication and power failures) (Chen et al., 2016; Nova et al., 2022), and the dynamic nature of parking environments (Martynova et al., 2024; Xiao et al., 2021). For that, more recent advancements have incorporated multimodal data fusion techniques to enhance prediction accuracy through integration of diverse data sources (Huang et al., 2024; H. Zhang et al., 2024). Furthermore, with the rise of smart parking guidance and information systems (PGIS) as well as smartphone applications, some research has shifted toward predicting parking behavior to guide drivers to vacant spaces more efficiently (Gao et al., 2021).

## 1.2  Related works

Methodologies for parking availability prediction can be systematically classified into knowledge-based and data-driven approaches (Huang et al., 2024; Xiao et al., 2023). Knowledge-based methods, which rely on extensive domain-specific assumptions and rule-based systems, demonstrate significant limitations in capturing the multifactorial complexity of parking dynamics in modern urban environments, particularly when confronted with nonlinear interactions between temporal patterns, spatial correlations, and external influencing factors (Martynova et al., 2024; Xiao et al., 2023). Whereas the currently dominant data-driven approaches have evolved through three phases: statistical methods, machine learning (ML), and deep learning (DL). Traditional statistical models such as autoregressive integrated moving average (ARIMA) and historical averaging (HA) demonstrate reasonable accuracy in short-term forecasts (Bock et al., 2020; Chawathe, 2019; Chen et al., 2013), but their performance deteriorates in long-term prediction due to inherent limitations in modeling complex temporal fluctuations and nonlinear system dynamics. To address these shortcomings, ML techniques such as decision trees, random forests, and support vector machines (SVMs) were introduced, offering improved adaptability to temporal variability and demand dynamics (Awan et al., 2020; Feng et al., 2019). However, conventional ML approaches suffer from several critical limitations, including extensive hyperparameter tuning requirements, computational inefficiency when handling large-scale datasets, and vulnerability to regional overfitting in spatially heterogeneous urban environments.

In recent years, the emergence of DL has revolutionized parking availability prediction and has been established as the prominent method. Recurrent neural network architectures, particularly Long Short Term Memory (LSTM) and Gated Recurrent Unit (GRU) variants, have demonstrated significant improvements in prediction accuracy when integrated with domain-specific optimization frameworks (Qiu et al., 2018; Rong et al., 2018; Shao et al., 2019). The Transformer architecture (Vaswani et al., 2017), with its attention mechanism and capacity for modeling long-range dependencies, has further revolutionized time-series forecasting. Building on the foundation of Transformer, subsequent research focuses on enhancing temporal correlation modeling. Typically, Informer (Zhou et al., 2021) achieves efficient long-sequence processing through ProbSparse self-attention mechanisms, while Autoformer (Wu et al., 2021) optimizes long-term forecasting via temporal decomposition architectures and Auto-Correlation mechanisms. These Transformer variants have demonstrated substantial success in transportation research, particularly in applications such as traffic speed prediction (Xue et al., 2025) and traffic state forecasting (Yu et al., 2024). Given their proven effectiveness, they are adopted in this study as benchmark models.

To address spatial-temporal correlations in parking systems, hybrid and sophisticated deep learning methods have been explored. For instance, Ghosal et al. (2019) developed a DL model that incorporates a convolutional neural network (CNN) and a stacked LSTM autoencoder for parking availability prediction. Recognizing that parking networks are inherently non-Euclidean, graph-based prediction algorithms, typically Graph Neural Network (GNN), have been introduced. Gong et al. (2021) employed a spatial-aware Graph Convolutional Network (GCN) for parking availability analysis and forecasting before and during the COVID-19 pandemic, while Xiao et al. (2021) developed a Hybrid Spatial-Temporal GCN (HST-GCN), further enhancing temporal modeling, specifically for on-street parking prediction. Zhang et al. (2022) proposed a Semi-supervised Hierarchical Recurrent Graph Neural Network-X (SHARE-X), which integrates a hierarchical graph convolution module to capture non-Euclidean spatial autocorrelations among parking lots, alongside contextual and multi-resolution soft clustering convolution blocks to model both local and global spatial dependencies. Wang et al. (2025) introduced an advanced graph coarsening framework that dynamically constructs parking graphs by leveraging real-time parking service capabilities. This approach significantly enhances both prediction accuracy and computational efficiency when applied to large-scale parking data, addressing critical scalability challenges inherent in traditional graph-based methods. Despite these advances, GNN-based approaches still face notable challenges in parking availability prediction. Their high computational complexity is poorly suited to the dynamic spatio-temporal volatility of parking data and its pronounced local characteristics. Moreover, reliance on explicitly defined graph structures imposes scalability constraints, particularly when processing large-scale heterogeneous multimodal transportation datasets from different sources.

Contemporary research has expanded to incorporate diverse contextual factors, including weather conditions, holiday effects, and spatial zoning characteristics (Xiao et al., 2023; Yang et al., 2019; H. Zhang et al., 2024). Nevertheless, existing models continue to struggle with capturing the inherent instability of parking dynamics, which exhibit high volatility due to factors such as parking duration uncertainty and demand fluctuations (Jiang et al., 2025; Xie et al., 2024). This challenge has motivated the development of advanced feature interaction modeling strategies for multivariate time-series, where recent approaches diverge into two distinct paradigms: Channel-independent models, such as DLinear (Zeng et al., 2023) and patch time series Transformer (PatchTST) (Nie et al., 2023), generally process each feature channel separately (via per-channel linear projections, decomposition, or patching) to mitigate noise propagation and reduce learning complexity; while channel-dependent models such as Crossformer (Zhang and Yan, 2023) and Inverted Transformer (iTransformer) (Liu et al., 2023) explicitly learn cross-variable interactions through attention mechanisms applied across feature dimensions (or by inverting the time/channel axes), enabling the discovery of latent inter-channel dependencies. Each approach has trade-offs: channel-independent designs are often more robust to noisy or weak channels and are computationally lighter, while channel-dependent architectures can capture richer cross-modal signals but demand more data, computation, and careful regularization. Despite these advances, critical knowledge gaps persist regarding which auxiliary features yield optimal predictive benefits and what feature fusion architectures best enhance prediction accuracy for parking availability forecasting tasks characterized by high spatio-temporal volatility. Motivated by these gaps, this study introduces a Self-supervised learning enhanced Spatio-temporal Inverted Transformer (SST-iTransformer) that integrates multi-source heterogeneous demand data via a dual-branch attention mechanism, aiming to advance the state of the art in parking availability prediction while maintaining computational feasibility for large-scale urban applications.

### 1.3 Contribution of this study

This study makes significant contributions to the field of parking availability prediction by addressing critical gaps identified in the literature review. Specifically, this study introduces a method that considers spatial-temporal correlations among parking lots with similar characteristics by constructing Parking Cluster Zones (PCZs) using K-means clustering. Within each PCZ's radiation area, comprehensive demand-related feature representations are constructed by integrating multi-source demand-related travel records from multiple transportation modes, including metro, bus, online ride-hailing, and taxi services.

To leverage these features, the core innovation of this work is the development of the SST-iTransformer framework, which fundamentally integrates the masking-reconstruction-based pretext tasks for self-supervised spatio-temporal representation learning and rethinks the attention mechanism design for parking availability prediction tasks. Unlike conventional Transformer variants, the proposed architecture features a dual-branch attention mechanism specifically engineered for high-volatility parking dynamics: Series Attention captures long-range temporal dependencies through patch-based operations, while Channel Attention models cross-variate interactions through inverted dimensions.

To rigorously validate the proposed approach, comprehensive experiments were conducted using real-world empirical data from Chengdu, China, comparing SST-iTransformer against an extensive set of baseline models, including traditional DL models (e.g., GRU, LSTM), Transformer variants (e.g., Informer, Autoformer), channel-independent architectures (e.g., DLinear, PatchTST), and channel-dependent models (e.g., Crossformer, iTransformer). The results demonstrate that the proposed framework consistently outperforms state-of-the-art benchmarks across short-term, medium-term, and long-term prediction horizons.

In summary, the main contributions of this paper lie in:

- **Multi-source heterogeneous mobility data fusion framework**: A comprehensive approach for integrating diverse transportation data sources (metro, bus, ride-hailing, taxi) was developed and validated to enhance parking availability prediction. Rigorous ablation studies quantitatively evaluated the relative importance of different data sources, revealing that ride-hailing data contributes most significantly to prediction accuracy with the largest performance degradation when omitted, followed by taxi data, while fixed-route transit features show marginal effects.
- **Self-supervised learning enhanced Spatio-temporal Inverted Transformer (SST-iTransformer)**: A novel Transformer variant, SST-iTransformer, enhanced by self-supervised spatio-temporal representation learning, was proposed. The SST-iTransformer architecture overcomes limitations of conventional approaches by simultaneously modeling long-range temporal dependencies and cross-variate interactions through Series Attention and Channel Attention, thereby achieving superior performance. The incorporation of self-supervised pretraining, utilizing masking-reconstruction-based pretext tasks for spatio-temporal representation learning, helps the model to uncover latent spatio-temporal patterns and dependencies, further enhancing its overall efficacy.
- **Comprehensive empirical validation**: Rigorous quantitative benchmarking was conducted using real-world datasets, evaluating model performance across varying prediction horizons and input sequence lengths. The framework achieved state-of-the-art results with the best mean squared error (MSE), outperforming existing methods.

## 2. Methodology

### 2.1 Problem formulation

Parking availability refers to the number of vacant parking spaces in a facility. For a system comprising $M$ parking lots, the system-wide parking availability state at time $t$ can be expressed as $Y^t = \{y_1^t, y_2^t, y_3^t, \cdots, y_M^t\} \in \mathbb{R}^M$, with $y_i^t$ denoting the number of available spaces in the parking lot $i$ at time $t$. Concurrently, the associated exogenous variables influencing parking dynamics at time $t$ can be noted by $X^t \in \mathbb{R}^{M \times D}$, with $D$ representing the dimensionality of the integrated feature set.

Given the historical parking availability states $Y^{t-L:t} \in \mathbb{R}^{M \times L \times 1}$ and corresponding exogenous features $X^{t-L:t} \in \mathbb{R}^{M \times L \times D}$ over $L$ time steps, the objective is to learn a mapping function $f(\cdot)$ to predict parking availability over the subsequent future $T$ steps, expressed as:

$$Y^{t+1:t+T} = f([Y^{t-L:t}, X^{t-L:t}]) \qquad (1)$$

where $Y^{t+1:t+T} \in \mathbb{R}^{M \times T \times 1}$ represents the forecasted parking availability states, $Y^{t-L:t}$ encodes temporal attributes of historical parking availabilities, and $X^{t-L:t}$ integrates additional exogenous variables, including, for example, weather conditions, traffic status, and demand patterns from other transportation modes (e.g., taxi, ride-hailing).

## 2.2 Demanding features integration

### 2.2.1 Parking cluster zone (PCZ)

The PCZ is defined as the area where travelers engage in various trip-related activities after parking their vehicles. The dynamic behavior of vehicles entering and exiting a parking lot is closely associated with the demand characteristics of the surrounding area, and the parking lots within the same vicinity often exhibit similar or correlated availability patterns. To capture these correlations and investigate parking availability more efficiently, this study utilized K-means clustering to group parking lots with comparable characteristics within designated areas together into the same PCZ. Considering the layout of urban streets, this study employed the Minkowski distance to compute the necessary buffer zone and then applied spatial concatenation to construct the corresponding PCZs.

As illustrated in **Fig. 1**, each "P" point represents a parking lot; different colors represent distinct parking lot clusters; "▼" denotes the cluster centroid. Four representative examples of PCZs are illustrated by the irregular polygons with black solid boundary lines encompassing relevant parking lots. It is noteworthy that PCZs vary in shape and in the number of contained parking lots, which reflects the heterogeneous spatial distribution of urban parking facilities and aligns with the underlying street layout and functional areas.

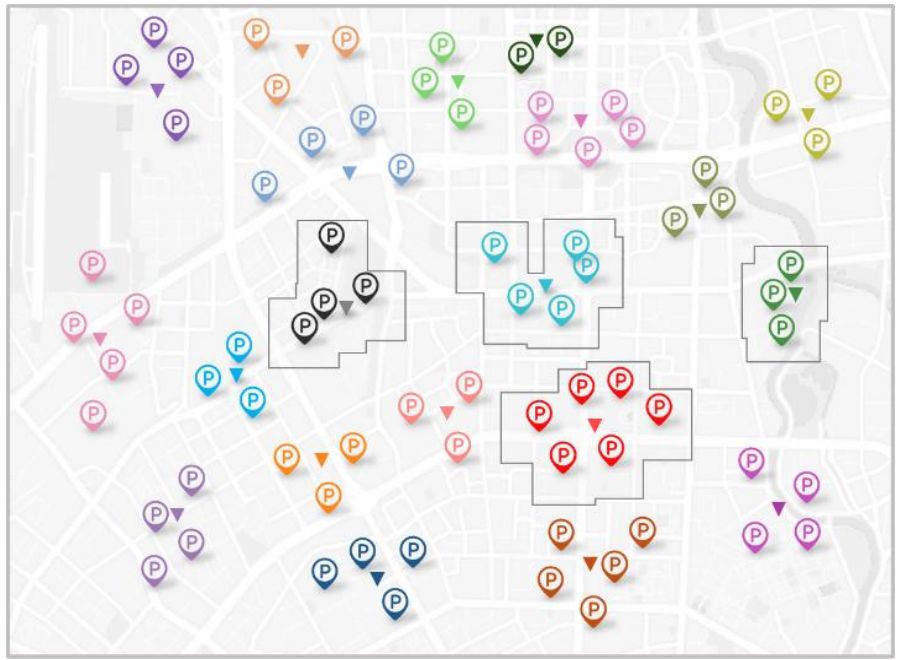

**Fig. 1.** The distribution of the parking lots and parking cluster zones (PCZs)

### 2.2.2 Spatial-temporal demanding feature integration

With the advancement of comprehensive transportation integration and the widespread adoption of Internet of Things (IoT) and information and communications technologies (ICT), data-driven fusion across heterogeneous travel modes has emerged as an effective approach for capturing and estimating dynamic demand patterns. Given their functional substitutability with private car usage, this study incorporates four representative modes, i.e., metro, bus, online ride-hailing, and taxi, for spatial-temporal demand feature fusion and forms an integrated multimodal demand profile. The detailed procedure and methodology for generating this integrated multimodal demand profile are formalized in **Algorithm 1** with pseudo code and conceptually illustrated in **Fig. 2.**

As delineated in **Algorithm 1** and **Fig. 2**, trip data from these four modes are processed to identify their corresponding origins (O) and destinations (D). A trip is associated with a PCZ if either its origin or destination lies within the PCZ boundary. The aggregated multimodal trip volumes are then fused to characterize the spatio-temporal demand profile of each PCZ. It is important to note that, due to local operational regulations (where card swipes are not required upon alighting), bus trip records in the study area include only boarding information (origins) without destinations. Accordingly, only origins are considered for the bus mode. Furthermore, to be comparable across modes and with other features (e.g., the historical parking availability), the demand values are normalized for each mode prior to integration. This normalization and integration process provides a unified multimodal demand feature, thereby enhancing the robustness and interpretability of subsequent model training.

---

**Algorithm 1: Integrated Multimodal Demand Calculation**

---

**Inputs:**
- *trip_data*: records of individual trips (mode, origin, destination, timestamp, etc.)
- *PCZ boundaries*: spatial boundaries of parking cluster zones (PCZs)

**Outputs:**
- *Integrated_Norm_Demand [PCZ, t]*: normalized integrated demand for each PCZ at time $t$

---

*Step 1: Initialization*

Initialize **Integrated_Demand**$[PCZ, t] \leftarrow 0$
Initialize **Count**$[mode, PCZ, t] \leftarrow 0$

---

*Step 2: Trip assignment*

**For** each *trip* in *trip_data* **do**:
   **Extract** *origin*, *destination*, $t$ = timestamp, *mode* // *mode* ∈ {bus, metro, taxi, ridehailing}
   **If** (*mode* == bus):
     **For** each *PCZ* in *PCZs* **do**:
       **If** *origin* is within the *PCZ*'s boundary:
         Count[*mode*, *PCZ*, *t*] = Count[*mode*, *PCZ*, *t*] + 1
       **End If**
     **End For**
   **Else**:
     **For** each *PCZ* in **PCZs Do**:
       **If** *origin* is within the *PCZ*'s boundary:
         Count[*mode*, *PCZ*, *t*] = Count[*mode*, *PCZ*, *t*] + 1
       **Else if** *destination* is within the *PCZ*'s boundary:
         Count[*mode*, *PCZ*, *t*] = Count[*mode*, *PCZ*, *t*] + 1
       **End If**
     **End For**
   **End If**
**End For**

---

*Step 3: Normalization and integration*

**For** each *mode* ∈ {bus, metro, taxi, ridehailing} and at each *PCZ* **do**:
   //norm the trip number along the *time* axis
   Norm_Demand[*mode*, *PCZ*, *t*] = Normalize(Count[*mode*, *PCZ*, *t*])
**End For**

**For** each *PCZ* at each time *t* **do**:
   //integrate the normalized demand among all modes
   Integrated_Norm_Demand[*PCZ*, *t*] = $\sum_{mode}$ Norm_Demand[*mode*, *PCZ*, *t*]
**End For**

---

**Output:** Integrated_Norm_Demand[*PCZ*, *t*]

---

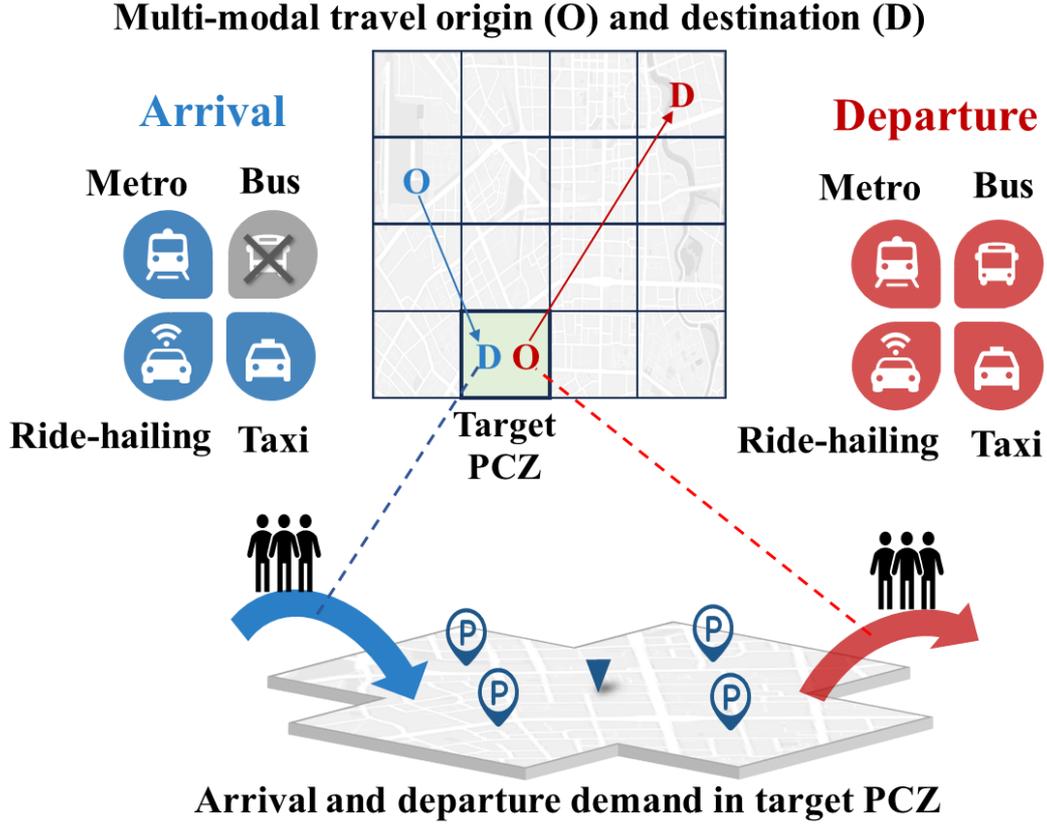

**Fig. 2.** Fusion of demand characteristics within parking cluster zone (PCZ)

The integrated spatio-temporal demand profile is defined at the PCZ level, meaning that all parking lots within a given PCZ share the same value of this integrated feature. This design captures collective demand patterns within each PCZ rather than treating parking lots in isolation, reflecting localized mobility interactions more effectively.

In summary, the proposed clustering-based approach generates PCZs with coherent boundaries that group spatially correlated parking lots together. Plus, the integrated multimodal demand profile provides a robust representation of arrival and departure demand within each PCZ. Consequently, the constructed input features, i.e., $[Y^{t-L:t}, X^{t-L:t}]$ in **Equation (1)**, captures both spatial correlations and temporal variations as well as collective demand patterns associated with parking dynamics, thereby enhancing the effectiveness of parking availability prediction.

### 2.3 Self-supervised learning enhanced Spatio-Temporal Inverted Transformer

#### 2.3.1 Overall framework

**Fig. 3** presents the overall architecture of the proposed Self-supervised learning enhanced Spatio-Temporal Inverted Transformer (SST-iTransformer), designed for parking availability prediction. The framework adopts an encoder–decoder structure with a novel masking-based self-supervised strategy to capture spatio-temporal dependencies effectively.

Raw input data first undergoes temporal feature encoding and is then transferred to the encoder. In the encoder, dual-branch feature extraction modules operate in parallel: Series dependency learning employs patch embedding and series attention to capture long-range temporal patterns; Channel-wise correlation learning models auxiliary feature interactions across spatial dimensions through channel embedding and channel attention. The outputs of both branches (independently processed features) are subsequently reshaped and reintegrated to generate the encoder output.

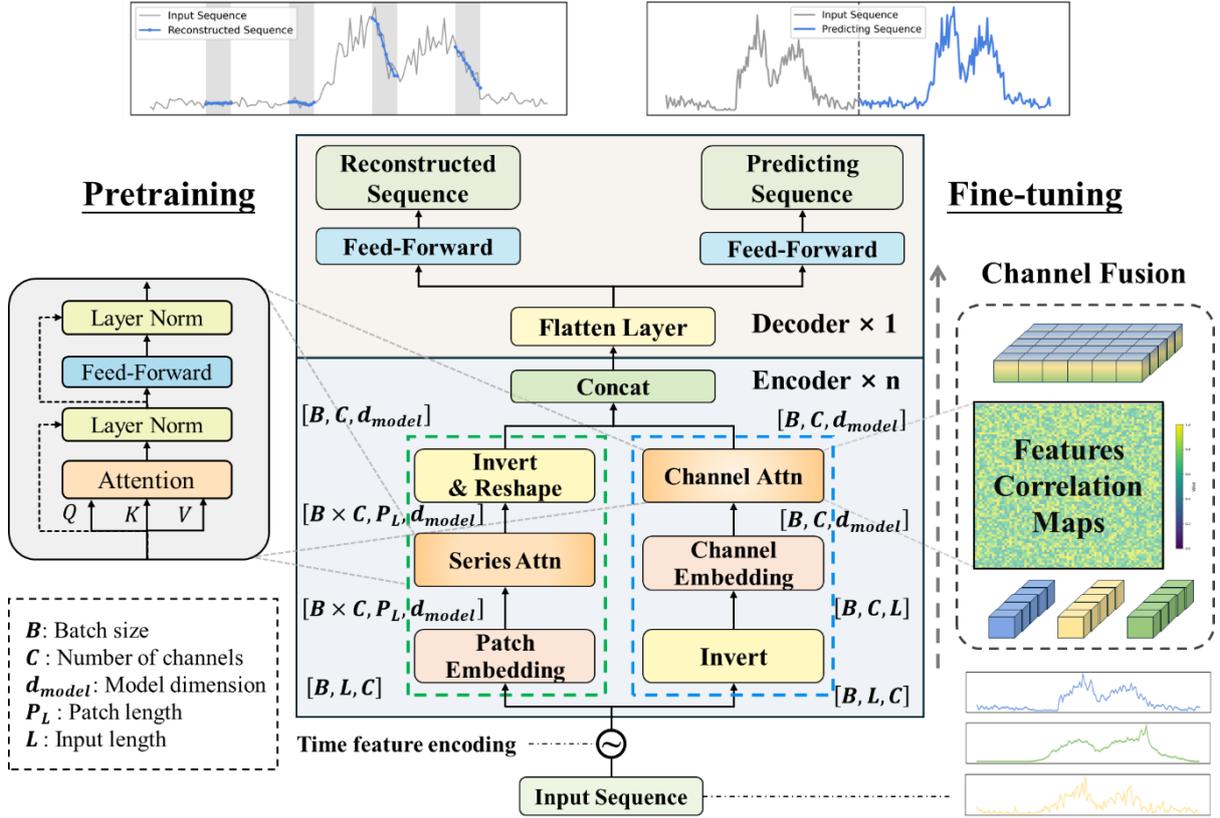

**Fig. 3.** The architecture of the self-supervised learning enhanced Spatio-temporal inverted Transformer

Transitioning to the decoding phase, in the decoder, specialized feed-forward networks adapt to task objectives. During pretraining, the model reconstructs masked patch sequences to enforce the learning of contextual spatio-temporal representations. During fine-tuning, it produces probabilistic multi-horizon predictions tailored for parking availability prediction. This integrated architecture enables robust self-supervised representation learning via masking-reconstruction-based pretext tasks, while maintaining task-specific optimization capabilities for prediction.

### 2.3.2 Self-Supervised learning strategy

Self-supervised learning has emerged as a transformative paradigm initially pioneered in representation learning to leverage intrinsic data structures without human annotations, notably advanced by masked autoencoder architectures in natural language processing (Devlin et al., 2019) and computer vision (Dong et al., 2025; He et al., 2022; Li and Dong, 2023). Instead of relying on costly labeled data, self-supervised learning constructs supervisory signals through pretext tasks (e.g., masking part of the images or texts and reconstructing them) on unannotated datasets.

In the context of parking availability prediction, self-supervised learning for model pretraining can offer distinct advantages: it alleviates data scarcity and missing values, enhances model generalization by learning robust spatio-temporal representations from abundant unlabeled data, and helps the model to excavate latent patterns such as temporal periodicity and spatial interdependencies that are often overlooked or hard to capture in fully supervised models.

In the proposed framework, reconstructing the deliberately masked spatio-temporal data in selected regions (gray areas in **Fig. 3**) serves as the pretext task in the pretraining phase. By masking selected continuous temporal segments (temporal masking) and selected spatial nodes (spatial masking), the model is forced to infer missing information to learn occluded patterns from the broader global context rather than simple local interpolation. This reconstruction is done by minimizing the MSE between the model prediction and the ground truth of the masked spatio-temporal data through backpropagation.

This masking-reconstruction-based self-supervised learning paradigm strengthens the model's capacity to capture complex parking occupancy dependencies, promoting the learning of robust and generalizable spatio-temporal representations.

### 2.3.3 Attention mechanism

**(1) Vanilla attention**

The attention mechanism, a cornerstone of modern DL architectures, dynamically weights the relevance of input elements when generating outputs in sequence-to-sequence modeling (Vaswani et al., 2017). By selectively focusing on pertinent segments of the input sequence during each decoding step, the attention mechanism overcomes the fixed-context limitations of CNNs and RNNs, enabling explicit modeling of long-range dependencies and capturing relationships between distant elements more effectively. First introduced in the Transformer architecture (Vaswani et al., 2017), attention has demonstrated significant cross-domain efficacy in natural language processing, computer vision, and long-term time-series forecasting. Its capacity to allocate contextually adaptive computational resources to critical input regions fundamentally enhances representational learning for complex spatio-temporal patterns, as evidenced by state-of-the-art performance in tasks requiring nuanced contextual understanding.

With Query-Key-Value (QKV) interior design, the scaled dot-product attention used in Transformer (Vaswani et al., 2017) is given by:

$$\text{Attention}(Q, K, V) = \text{softmax}(\frac{QK^T}{\sqrt{d_{model}}})V \quad (2)$$

where $Q$, $K$, and $V$ are query, key, and value vectors, respectively, and $d_{model}$ denotes the dimensionality of the model. $Q$, $K$, and $V$ are calculated by linearly converting the input sequence.

**(2) Series attention**

In the proposed SST-iTransformer, the Series Attention module employs a patching operation to segment the input time series tensor $Z \in \mathbb{R}^{B \times L \times C}$ into local subseries along the temporal dimension. Here, $B$ is the batch size, $L$ is the sequence length, and $C$ is the number of channels (i.e., number of feature variates). For each variate $c$ and batch element $b$, the univariate time series $Z_{b,:,c} \in \mathbb{R}^L$ is partitioned into $\text{patch\_num} = [L/P_L]$ non-overlapping patches, where $P_L$ is the patch length. The patching operation can be formalized by the following equations. For a single patch indexed by $p \in \{1, 2, \ldots, \text{patch\_num}\}$:

$$z_{b,c}^{(p)} = Z_{b,(p-1) \times P_L + 1 : p \times P_L, c} \in \mathbb{R}^{P_L} \quad (3)$$

$z_{b,c}^{patch}$ stacks all patches across the sequence:

$$z_{b,c}^{patch} = [z_{b,c}^{(1)}, z_{b,c}^{(2)}, \ldots, z_{b,c}^{(p)}, \ldots, z_{b,c}^{(patch\_num)}] \in \mathbb{R}^{patch\_num \times P_L} \quad (4)$$

Extending this operation across all batches and channels produces $Z^{patch} \in \mathbb{R}^{B \times patch\_num \times P_L \times C}$. The patching process segments the time series into smaller subseries (patches) to reduce sequence length and computational complexity, while enhancing the model's ability to capture localized temporal semantics.

Subsequently, the patched representation $Z^{patch}$ is passed through the Series Attention block to calculate the sequence attention scores through the following formula:

$$Z_{new}^{patch} = \text{SAttention}(Q_s, K_s, V_s) \quad (5)$$

where $Q_s = Z^{patch} W_{qs}$, $K_s = Z^{patch} W_{ks}$, and $V_s = Z^{patch} W_{vs}$; $W_{qs}$, $W_{ks}$, and $W_{vs}$ are learnable projection matrices.

**(3) Channel attention**

Channel Attention enhances the model's ability to emphasize important feature channels by adaptively rescaling their contributions. Unlike conventional Transformers, in the proposed SST-iTransformer, prior to entering the Channel Attention block, the data will undergo an invert operation:

$$Z_{in} = \text{Invert}(Z) \tag{6}$$

where $Z \in \mathbb{R}^{B \times L \times C}$ ($B$ is the batch size, $L$ is the sequence length, and $C$ is the number of channels) is transformed by the invert operation into $Z_{in} \in \mathbb{R}^{B \times C \times L}$, effectively permuting the dimensions. The invert operation maps the time series dimension onto a fixed hidden representation, thereby not only enhancing the model's adaptability to input sequences of varying lengths but also improving computational efficiency. It is also worth noting that, as shown in **Fig. 3**, the Series Attention branch also incorporates an invert-and-reshape step prior to its output, ensuring dimensional alignment with the Channel Attention branch and thereby enabling seamless concatenation.

After the invert operation, Channel Attention is computed using the following formula:

$$Z_{new} = \text{CAttention}(Q_c, K_c, V_c) \tag{7}$$

where $Q_c = Z_{in} W_{qc}$, $K_c = Z_{in} W_{kc}$, and $V_c = Z_{in} W_{vc}$; $W_{qc}$, $W_{kc}$, and $W_{vc}$ are learnable matrices.

**(4) Multi-head attention**

In multi-head attention, the query, key, and value matrices are divided and projected into multiple heads, each with its own set of parameters. Then, attention is computed independently for each head, and the results are concatenated and linearized.

With $h$ attention heads, the multi-head attention mechanism is formulated as follows:

$$\text{MHAttn}(Q, K, V) = \text{Concat}(head_1, \cdots, head_H) W^O \tag{8}$$

where $head_i = \text{Attention}(Q W_i^Q, K W_i^K, V W_i^V)$, and $W_i^Q, W_i^K, W_i^V$, and $W^O$ are learnable matrices.

### 2.3.4 Implementation details

**(1) Time feature encoding**

Positional encoding (PE) is a fundamental and crucial component in Transformer architectures, designed to compensate for the absence of inherent sequential information in time series data. In time series forecasting tasks, encoding temporal information is particularly critical for capturing inherent (e.g., periodic) patterns and dependencies. Timestamp encoding represents converting timestamps into a format that can be effectively used by ML models. One common method is to use *sine* and *cosine* functions to encode time of day, day of the week, month, etc. With "timestamp" as its associated timestamp value and "T" representing the maximum value of the time feature (e.g., maximum hour in a day, maximum day in a week), the timestamp encoding can be computed as follows:

$$PE_{\sin}(timestamp, f) = \sin\left(timestamp \times \frac{2\pi}{T}\right) \tag{9}$$

$$PE_{\cos}(timestamp, f) = \cos\left(timestamp \times \frac{2\pi}{T}\right) \tag{10}$$

where $f \in \{hour, day, week, month\}$ represents the periodic feature of the input embeddings.

**(2) Loss functions**

The SST-iTransformer employs dual objective functions across training phases. During pretraining, a self-supervised reconstruction loss is computed as MSE:

$$\mathcal{L}_{\text{reconstruction}} = \frac{1}{N_{\text{mask}}} \sum_{n=1}^{N_{\text{mask}}} ([x,y]_{masked} - [x,y]_{reconstructed})^2 \tag{11}$$

where $[x,y]_{masked}$ denotes the ground-truth values of the masked regions, which incorporate both the parking availability state $y$ and the associated exogenous feature $x$, and $[x,y]_{reconstructed}$ represents the reconstructed outputs by model learning and estimation.

For fine-tuning, the primary loss measures prediction error against ground-truth parking availability using:

$$\mathcal{L}_{\text{forecast}} = \frac{1}{N} \sum_{n=1}^{N} \left(y_{true}^{t+1:t+T} - y_{output}^{t+1:t+T}\right)^2 \tag{12}$$

where $y_{true}^{t+1:t+T}$ and $y_{output}^{t+1:t+T}$ represent the true values and predicted values, respectively, over the prediction horizon $T$ time steps.

### (3) Layer normalization

Layer Normalization (LayerNorm) is a technique used to stabilize and accelerate the training of deep neural networks by normalizing the inputs across the feature dimension for each sample. To avoid notation conflict with time series data $x$ used in previous equations, the input to LayerNorm is denoted as $h$, then the LayerNorm process is defined as:

$$\overline{h_i} = \frac{h_i - \mu}{\sqrt{\sigma^2 + \varepsilon}} \cdot \gamma + \beta \tag{13}$$

where $h_i$ is the input value, $\mu$ is the mean, $\sigma^2$ is the variance computed over the feature dimension, $\varepsilon$ is a small constant for numerical stability; $\gamma$ and $\beta$ are learnable scaling and shifting parameters.

### (4) Feed-forward network

In the Transformer architecture, the Feed-forward Neural Network (FNN) is connected to each attention layer, serving as a critical component for nonlinear feature transformation. The FNN is a fully connected module, defined as

$$\text{FNN}(H') = \text{ReLU}(H'W_1 + b_1)W_2 + b_2 \tag{14}$$

where $H'$ is outputs of the previous layer, $W_1 \in \mathbb{R}^{d_{model} \times 4d_{model}}$ and $W_2 \in \mathbb{R}^{d_{model} \times 4d_{model}}$ denote trainable weight matrices, $b_1 \in \mathbb{R}^{4d_{model}}$ and $b_2 \in \mathbb{R}^{4d_{model}}$ represent bias vectors. Here, $d_{model}$ denotes the model's hidden dimension (consistent with the architectural specifications in **Fig. 3**).

The intermediate expansion to $4d_{model}$ dimensions enhances the network's capacity to learn complex representations, while the ReLU activation introduces nonlinearity. Together, these mechanisms enable the model to capture higher-order spatio-temporal dependencies that are critical for accurate parking availability prediction.

## 3. Experiments and results comparison

### 3.1 Data description and preprocessing

This study utilized parking records from 168 parking lots in Chengdu, Sichuan Province, China. The dataset encompasses detailed vehicle arrival times, departure times, dates, and total parking capacity for each parking spot. Data collection spanned a full month, from September 1 to September 30, 2021, comprising a total of 3,803,570 parking records. To capture spatial correlations, the 168 parking lots were systematically clustered into 30 Parking Cluster Zones (PCZs) using K-means clustering based on factors such as spatial distribution and traffic in/outflow patterns. Additionally, multimodal transportation data (including metro, bus, online ride-hailing, and taxi trip records) within these 30 PCZs were integrated to construct comprehensive demand feature representations.

For data preprocessing, a rigorous multi-step refinement strategy was implemented to enhance data quality and mitigate noise and inconsistencies. Initially, records containing missing or NULL values were systematically eliminated using the *dropna* function in Python. Subsequently, the data was aggregated into 10-minute intervals to reduce temporal noise while preserving meaningful patterns. To mitigate the impact of low-frequency noise components, a discrete Fourier Transform was applied to convert the time-domain data to the frequency domain, followed by spectral analysis, selective filtering to attenuate noise components beyond the dominant signal bandwidth, and an inverse Fourier Transform to reconstruct the denoised time-series data. After comprehensive preprocessing, 4,320 refined temporal records were obtained per parking lot, yielding a total of 725,760 samples (4,320 × 168 parking lots).

### 3.2    Experimental configuration

The processed dataset was partitioned into training (60%), validation (10%), and testing (30%) subsets through stratified temporal sampling to maintain temporal continuity while ensuring representative distribution of operational conditions. The experimental design implemented a sequence-to-sequence prediction framework wherein historical observations from the preceding 144 time steps (equivalent to 24 hours of data at 10-minute resolution) served as input to forecast parking availability across the subsequent 144 time steps (spanning 10 minutes to 24 hours into the future). This multi-horizon prediction framework explicitly addresses both short-term operational needs (e.g., immediate parking guidance) and medium-term planning requirements (e.g., resource allocation for the coming day), reflecting practical deployment scenarios in intelligent parking management systems.

During model training, data from all 168 parking lots were utilized to maximize spatial coverage and capture heterogeneous patterns. For the testing phase, a rigorous out-of-sample validation protocol was employed using data from 72 parking lots distributed across 10 distinct PCZs, ensuring comprehensive evaluation across diverse spatial contexts while maintaining independence from the training set. This experimental configuration provides a robust assessment of model generalizability across both temporal and spatial dimensions, a critical requirement for real-world deployment in dynamic urban environments.

### 3.3    Baseline models

This study evaluates the performance of the proposed SST-iTransformer against a comprehensive set of baseline models spanning three methodological categories: traditional RNNs (including GRU and LSTM), Transformer variants with enhanced temporal modeling capabilities (e.g., Informer, Autoformer), as well as specialized architectures for channel-independent (e.g., DLinear, PatchTST), and channel-dependent cross-variate dependency modeling (e.g., Crossformer, iTransformer).

- **Gated Recurrent Unit model (GRU):** GRU is a RNN variant that efficiently captures long-term dependencies in sequential data via a gating mechanism, which employs update and reset gates to selectively regulate information flow, mitigating issues like vanishing gradients and enhancing the model's ability to learn complex temporal patterns (Cho et al., 2014).
- **Long Short-Term Memory (LSTM):** LSTM is another RNN variant designed to handle the problem of long-term dependency and mitigate the issue of vanishing gradients (Hochreiter and Schmidhuber, 1997). LSTM processes sequential data and outputs information for a specific window of time steps. A typical LSTM unit is composed of a cell, an input gate, an output gate, and a forget gate.
- **Standard Transformer:** Transformer is a deep neural network architecture based entirely on self-attention mechanisms, eliminating recurrence to capture long-range dependencies in sequential data (Vaswani et al., 2017). It processes entire sequences in parallel, making it highly efficient for time series forecasting. In this study, a standard Transformer encoder-decoder structure is adopted.
- **Informer:** Informer is a Transformer variant for long-sequence time-series forecasting. It employs ProbSparse self-attention to reduce computational complexity and distillation operations to prioritize dominant features (Zhou et al., 2021).
- **Autoformer:** Autoformer is a Transformer variant incorporating a series decomposition architecture integrated with an auto-correlation mechanism. It automatically decomposes time series into trend and seasonal components and replaces point-wise self-attention with sub-series-level dependency aggregation based on periodicity (Wu et al., 2021).
- **DLinear:** DLinear is a simple yet powerful linear model that explicitly applies series decomposition followed by channel-independent linear projections (Zeng et al., 2023). Its emphasis on direct temporal modeling establishes it as a critical baseline to evaluate whether complex DL models genuinely outperform linear approaches for parking prediction.
- **PatchTST:** PatchTST is a Transformer-based model for multivariate time series forecasting that introduces patching of input sequences and channel-independent modeling. By segmenting time series into subseries-level patches as tokens, it captures local temporal patterns and reduces computational complexity while maintaining global dependencies via self-attention (Nie et al., 2023).
- **Crossformer:** Crossformer is a Transformer variant that focuses on capturing multi-scale temporal dependencies and cross-variable interactions in multivariate time series. Its core innovation includes two-stage attention (dimension-segment-wise and time-segment-wise) and a hierarchical encoder-decoder that fuses information across time scales and variates (Zhang and Yan, 2023).
- **iTransformer:** iTransformer is a novel Transformer variant that fundamentally rethinks traditional architecture by treating time points as tokens and variates as sequences. It employs channel-independent embedding, layer normalization, and applies self-attention across variate dimensions (rather than time steps) to better capture cross-variate dependencies and temporal patterns (Liu et al., 2023).

## 3.4 Evaluation metric

To comprehensively evaluate model performance for parking availability prediction, this study employs three complementary metrics:

- Mean Squared Error (MSE):

$$\text{MSE} = \frac{1}{N_{\text{test}}} \sum_{n=1}^{N_{\text{test}}} (\hat{y}_n - y_n)^2 \qquad (15)$$

- Mean Absolute Error (MAE):

$$\text{MAE} = \frac{1}{N_{\text{test}}} \sum_{n=1}^{N_{\text{test}}} |\hat{y}_n - y_n| \qquad (16)$$

- Mean Absolute Percentage Error (MAPE):

$$\text{MAPE} = \frac{100\%}{N_{\text{test}}} \sum_{n=1}^{N_{\text{test}}} \left| \frac{\hat{y}_n - y_n}{y_n} \right| \qquad (17)$$

Here, $N_{\text{test}}$ signifies the total number of test samples, while $y_n$ and $\hat{y}_n$ represent the actual and predicted values, respectively. MSE, MAE, and MAPE are three prevalent metrics for predictive analysis. MSE emphasizes larger errors through squaring, and is thus appropriate for penalizing larger errors; MAE takes the absolute value of the error, providing a more robust measure that reflects the average level of actual error; MAPE normalizes errors relative to actual values, enabling comparison across different scales, making it more feasible for prediction tasks of varying magnitudes.

Moreover, the model parameter size, represented as Params in millions (M), along with the multiply-accumulate operations, denoted as MACs in gigaflops (G), serve as indicators for estimating the computational complexity and real-time deployment feasibility.

## 3.5 Results comparison

**Table 1** summarizes the quantitative performance comparison across evaluated models, with optimal results highlighted in bold and second-best results underlined. Results represent mean values across the 72 target parking lots among the predicted 144 steps in the test dataset.

Table 1. Performance Comparison of the tested Models

| Algorithms | MSE | MAE | MAPE (%) | MACs(G) | Params(M) |
|---|---|---|---|---|---|
| GRU | 0.5923 | 0.4192 | 3.6040 | 3.5581 | 35.3646 |
| LSTM | 0.6004 | 0.4149 | 3.3947 | 3.6978 | 36.3374 |
| Transformer | 0.5294 | 0.3790 | 2.7005 | 3.4834 | <u>20.5158</u> |
| Informer | 0.6166 | 0.4204 | 3.2905 | 3.5747 | 24.6205 |
| Autoformer | 0.4494 | 0.4032 | 3.1732 | 3.5178 | 30.6689 |
| DLinear | 0.5916 | 0.4188 | 3.5118 | **0.0030** | **0.0418** |
| PatchTST | 0.3635 | **0.3494** | **2.2486** | 3.8766 | 40.7575 |
| Crossformer | 0.3596 | 0.3544 | <u>2.3634</u> | 3.5342 | 35.8303 |
| iTransformer | 0.3455 | 0.3577 | 2.7871 | <u>3.0135</u> | 32.2975 |
| iTransformer-patch | <u>0.3321</u> | 0.3546 | 2.8503 | 3.2247 | 37.5944 |
| SST-iTransformer | **0.3293** | <u>0.3523</u> | 2.8337 | 3.2247 | 37.5944 |

*Note: The best result is shown in **bold**, and the second-best result is <u>underlined</u>.*

As demonstrated in **Table 1**, traditional RNNs (i.e., GRU and LSTM) demonstrate limited effectiveness in capturing the complex patterns inherent in parking availability data, despite their sequential modeling capabilities. The standard Transformer model exhibits measurable improvement over these RNN-based approaches, confirming the advantage of its attention mechanism in capturing temporal dependencies. Autoformer further advances this performance through its innovative auto-correlation mechanism, representing a substantial enhancement over conventional attention formulations and demonstrating significant improvement over the baseline Transformer architecture.

Notably, Informer underperforms relative to both the standard Transformer and other contemporary approaches, a counterintuitive finding given its established efficacy in other time-series domains. This performance discrepancy may stem from the high volatility of parking dynamics, where Informer's ProbSparse attention mechanism, while computationally efficient, potentially omits critical short-term fluctuations essential for accurate parking prediction, and then degrades its performance below that of the conventional Transformer.

Channel-independent approaches yield mixed results. DLinear, despite its computational efficiency (the lightest in MACs(G) and Params(M)), fails to replicate its established effectiveness in parking availability prediction, potentially due to weaker associations within parking data compared to conventional time-series domains, limiting its decomposition approach to discern meaningful latent relationships. In contrast, PatchTST demonstrates significant enhancement over the traditional Transformer architecture, achieving the lowest absolute and relative errors (MAE = 0.3494, MAPE = 2.2486%).

Channel-dependent models show promising performance. Diverging from conventional approaches, Crossformer employs a two-stage attention mechanism to capture cross-temporal and cross-variate dependencies, yielding competitive mitigation of high-magnitude percentage errors with the second-best MAPE (2.3634 %). Similarly, the conventional iTransformer model also delivered strong performance, achieving competitive results with an MSE of 0.3455 and an MAE of 0.3577 while utilizing the minimal computational cost (measured in MACs). However, the conventional iTransformer still lightly overlooks the inherent temporal dependencies within sequential data.

To address this limitation, this study proposes two enhanced variants: iTransformer-patch, incorporating patch-based modifications, and SST-iTransformer, which further integrates a self-supervised training paradigm. Comparative analysis demonstrates that the refined iTransformer-patch model yields significant improvements over the conventional iTransformer across two key metrics: MSE and MAE. Further, the proposed self-supervised learning enhanced iTransformer-patch (SST-iTransformer) establishes state-of-the-art performance, achieving the lowest MSE (0.3293) alongside the second-best

MAE (0.3523), outperforming all other baselines, including the iTransformer-patch without self-supervised learning based pretraining. This performance superiority, attained with reasonable computational overhead, validates the efficacy of integrating self-supervised pretraining with the inverted Transformer's dimension-reversed architecture for capturing intricate spatio-temporal dependencies for parking availability forecasting.

## 4. Ablation studies

### 4.1 Multimodal transportation feature ablation analysis

#### 4.1.1 Transportation mode feature importance assessment

To systematically evaluate the contribution of heterogeneous transportation modalities to parking availability prediction, this study conducted comprehensive feature ablation experiments following a rigorous methodology adapted and upgraded from (Yang et al., 2022). The experimental design implemented five distinct feature configurations: (1) Full features, denoted by (**F**); (2) Full features excluding metro features (**F-M**); (3) Exclusion of bus features (**F-B**); (4) Exclusion of ride-hailing features (**F-R**); and (5) Exclusion of taxi features (**F-T**). These configurations were evaluated under two prediction paradigms: (a) predicting target parking availability using features from **all parking lots (APLs)** within PCZs; and (b) predicting parking availability of the **target parking lots (TPLs)** (i.e., the ones that need to be predicted) using exclusively the features from TPLs in PCZs.

Quantitative results, summarized in **Tables 2** and **3**, reveal significant heterogeneity in the predictive utility of different transportation modalities. Ride-hailing features demonstrate the most substantial contribution to prediction accuracy, with their exclusion (the columns under F-R) inducing the most severe performance degradation: the APL paradigm exhibits a mean MSE increase of 1.3991 ($\Delta$MAE = 0.4753), accumulated by all the tested models, while the TPL paradigm shows an MSE increase of 1.0611 ($\Delta$MAE = 0.4071). For the optimal model, SST-iTransformer, omitting demanding features from ride-hailing mode results in a 13.2% MSE increase under the APL paradigm, significantly surpassing other feature ablation scenarios. Taxi features represent the second most influential modality, with their exclusion (the columns under F-T) elevating MSE by 0.6047 (APL) and 0.4102 (TPL), indicating strong coupling between taxi demand and parking availability.

Conversely, excluding bus (the columns under F-B) or metro (the columns under F-M) features yields minimal performance degradation, particularly for metro features. This negligible impact suggests limited linkage between fixed-route transit dynamics and parking availability patterns.

The superior predictive utility of ride-hailing and taxi features can be attributed to their point-to-point service nature, which directly captures spatio-temporal correlations between individual mobility demand and parking behavior. In contrast, fixed-route public transport (bus/subway), constrained by predetermined routing and bulk passenger movement characteristics, displays weaker alignment with individual parking decisions.

#### 4.1.2 Spatial dependency analysis within PCZs

A comparative analysis of prediction paradigms through comparing results in **Tables 2** and **3** reveals the critical importance of spatial correlations among parking lots in the PCZs. Models relying solely on target parking lot characteristics (the TPL paradigm) consistently exhibit significantly higher prediction errors across all feature configurations compared to the APL paradigm. Under full-feature settings (F), SST-iTransformer demonstrates a 3.9% MSE increase in the TPL paradigm (0.3422) relative to the APL paradigm (0.3293). This performance divergence amplifies during feature ablation: excluding ride-hailing features (F-R) yields a $\Delta$MSE of 1.3991 under APL versus a $\Delta$MSE of 1.0611 under the TPL paradigm, confirming that data from correlated parking lots partially compensates for critical feature absence. These results unequivocally validate strong spatial dependencies among regional parking facilities, where neglecting cross-lot correlations with PCZs systematically degrades prediction performance.

Table 2. Model performance results for feature ablation comparison of the APL paradigm

| Models | F | | F-M | | F-B | | F-R | | F-T | |
|---|---|---|---|---|---|---|---|---|---|---|
| | MSE | MAE | MSE | MAE | MSE | MAE | MSE | MAE | MSE | MAE |
| GRU | 0.5923 | 0.4192 | 0.6328 | 0.4371 | 0.6372 | 0.4306 | 0.6884 | 0.4531 | 0.6913 | 0.4495 |
| LSTM | 0.6004 | 0.4149 | 0.6202 | 0.4106 | 0.7483 | 0.4663 | 1.0118 | 0.5685 | 0.6351 | 0.4262 |
| Transformer | 0.5294 | 0.3790 | 0.5314 | 0.4046 | 0.6315 | 0.4158 | 0.7235 | 0.4354 | 0.5826 | 0.3878 |
| Informer | 0.6166 | 0.4204 | 0.6812 | 0.4322 | 0.6724 | 0.4266 | 0.9185 | 0.5095 | 0.7678 | 0.4541 |
| Autoformer | 0.4494 | 0.4032 | 0.5619 | 0.4327 | 0.4940 | 0.3702 | 0.6964 | 0.4684 | 0.6000 | 0.4460 |
| DLinear | 0.5916 | 0.4188 | 0.5916 | 0.4188 | 0.5916 | 0.4188 | 0.5916 | 0.4188 | 0.5916 | 0.4188 |
| PatchTST | 0.3635 | 0.3494 | 0.3635 | 0.3494 | 0.3635 | 0.3494 | 0.3635 | 0.3494 | 0.3635 | 0.3494 |
| Crossformer | 0.3596 | 0.3544 | 0.3632 | 0.3364 | 0.3648 | 0.3497 | 0.3776 | 0.3424 | 0.3876 | 0.3544 |
| iTransformer | 0.3455 | 0.3577 | 0.3540 | 0.3673 | 0.3657 | 0.3599 | 0.3910 | 0.3749 | 0.3708 | 0.3740 |
| iTransformer-patch | 0.3321 | 0.3546 | 0.3371 | 0.3612 | 0.3543 | 0.3662 | 0.3738 | 0.3982 | 0.3631 | 0.3738 |
| SST-iTransformer | 0.3293 | 0.3523 | 0.3328 | 0.3570 | 0.3531 | 0.3521 | 0.3728 | 0.3806 | 0.3610 | 0.3742 |
| Δmetric | - | - | 0.2601 | 0.0835 | 0.4666 | 0.0816 | 1.3991 | 0.4753 | 0.6047 | 0.1843 |

*Note: The Δmetric is defined as the sum of the relative performance deviations across all tested models, with each deviation calculated against the full features (F) configuration.*

Table 3. Model performance results for feature ablation comparison of the TPL paradigm

| Models | F | | F-M | | F-B | | F-R | | F-T | |
|---|---|---|---|---|---|---|---|---|---|---|
| | MSE | MAE | MSE | MAE | MSE | MAE | MSE | MAE | MSE | MAE |
| GRU | 0.6378 | 0.4571 | 0.6429 | 0.4606 | 0.6876 | 0.4831 | 0.8676 | 0.5362 | 0.7152 | 0.4820 |
| LSTM | 0.6297 | 0.4482 | 0.6507 | 0.4531 | 0.6396 | 0.4632 | 1.0273 | 0.6103 | 0.6481 | 0.4517 |
| Transformer | 0.5892 | 0.4286 | 0.6413 | 0.4503 | 0.6180 | 0.4476 | 0.7217 | 0.4684 | 0.6933 | 0.4596 |
| Informer | 0.6648 | 0.4506 | 0.6849 | 0.4571 | 0.7166 | 0.4819 | 0.8321 | 0.5051 | 0.7658 | 0.4856 |
| Autoformer | 0.6639 | 0.4966 | 0.6326 | 0.5118 | 0.7178 | 0.4628 | 0.7158 | 0.5101 | 0.7297 | 0.5212 |
| DLinear | 0.5916 | 0.4188 | 0.5916 | 0.4188 | 0.5916 | 0.4188 | 0.5916 | 0.4188 | 0.5916 | 0.4188 |
| PatchTST | 0.3635 | 0.3494 | 0.3635 | 0.3494 | 0.3635 | 0.3494 | 0.3635 | 0.3494 | 0.3635 | 0.3494 |
| Crossformer | 0.3803 | 0.3682 | 0.3846 | 0.3562 | 0.3863 | 0.4000 | 0.3999 | 0.3626 | 0.3895 | 0.3753 |
| iTransformer | 0.3656 | 0.3897 | 0.3665 | 0.3759 | 0.3866 | 0.4024 | 0.3797 | 0.3960 | 0.3742 | 0.3862 |
| iTransformer-patch | 0.3560 | 0.3687 | 0.3584 | 0.3672 | 0.3645 | 0.3662 | 0.3728 | 0.3982 | 0.3631 | 0.3738 |
| SST-iTransformer | 0.3422 | 0.3527 | 0.3560 | 0.3570 | 0.3531 | 0.3521 | 0.3738 | 0.3806 | 0.3610 | 0.3742 |
| Δmetric | - | - | 0.0883 | 0.0289 | 0.2406 | 0.0989 | 1.0611 | 0.4071 | 0.4102 | 0.1292 |

*Note: The Δmetric is defined as the sum of the relative performance deviations across all tested models, with each deviation calculated against the full features (F) configuration.*

To further assess the impact of the proposed integrated demand features regarding spatial feature integration, systematic spatial feature ablation experiments were conducted. As shown in **Table 4**, four distinct settings were employed to evaluate the influence of different feature configurations:
(1) Comprehensive inclusion of all features (integrated spatio-temporal demand data and historical parking data from both target and related parking lots within PCZs);
(2) Integrated spatio-temporal demand data combined with historical parking data exclusively from the target parking lot;
(3) Historical parking data from both the target parking lot and related parking lots within the PCZs, paired with integrated spatio-temporal demand data only specific to the target parking lots;
(4) Historical data only from the target parking lot, supplemented by integrated spatio-temporal demand data and historical data from related parking lots within the PCZs.

**Fig. 4** compares the impact of incorporating spatio-temporal demand features and cross-parking-lot historical data on parking availability prediction performance (evaluated using MSE, MAE, and MAPE).

Table 4. Different feature configuration settings across PCZs

| Feature settings | Target parking lots | | Related parking lots within the PCZs | |
|---|---|---|---|---|
| | Historical parking data | Integrated demanding data | Historical parking data | Integrated demanding data |
| 1 | √ | √ | √ | √ |
| 2 | √ | √ | × | √ |
| 3 | √ | √ | √ | × |
| 4 | √ | × | √ | √ |

*Note: √ indicates "with", and × stands for "without" specific data.*

As illustrated in **Fig. 4**, setting 1 (integrating all features within PCZs) consistently yields optimal performance across all evaluation metrics (MSE, MAE, MAPE) for both baseline models (e.g., Transformer, Informer) and the proposed channel-relation enhanced variants (iTransformer-patch, SST-iTransformer). Comparisons on performance degradation in settings 2, 3, and 4 reveal that setting 2 exhibits the largest error increases, followed by setting 4, then setting 3. This hierarchy pattern indicates that historical data from related parking lots within PCZs constitutes the most effective auxiliary feature, while integrated demand data from the target lots provides secondary utility. Demand data from other related parking lots with the PCZ contributes minimally to prediction accuracy.

Notably, model sensitivity to feature ablation varies significantly across architectures. Conventional Transformer models (e.g., Transformer, Autoformer) suffer severe degradation under setting 2, suggesting their inherent linear layers implicitly capture spatial demand features despite lacking dedicated cross-channel mechanisms. In contrast, variants enhanced with patch-based sequence attention (iTransformer-patch and SST-iTransformer) demonstrate exceptional robustness to feature configuration changes. The baseline iTransformer, constrained by its channel-only attention mechanism, exhibits heightened susceptibility to feature availability variations, underscoring the importance of the proposed dual-branch attention design.

### 4.2 Comparative analysis of fine-tuning strategies and varying prediction horizons

**Table 5** presents a comprehensive evaluation of five different fine-tuning strategies across five varying prediction horizons ranging from short-term to extended long-term forecasting. Averaged testing results across the prediction windows spanning 24 hours (1-144 steps), 36 hours (1-216 steps), 48 hours (1-288 steps), 60 hours (1-360 steps), and 72 hours (1-432 steps) are illustrated. This multi-horizon prediction framework addresses diverse practical operational and strategic needs in intelligent parking management systems: short-term horizons support immediate operational requirements, such as real-time parking guidance for drivers; medium-term horizons facilitate daily planning, including resource allocation for the upcoming day; and longer horizons enable advanced strategic planning, such as multi-day resource optimization, staffing, infrastructure scheduling, and preparation for special events.

This systematic evaluation provides critical insights into the temporal robustness of different models and corresponding fine-tuning strategies of SST-iTransformer for parking availability prediction, revealing how architectural choices impact performance across varying forecasting durations.

As expected, all models exhibit increased prediction error as the forecasting horizon extends, consistent with theoretical expectations of error accumulation in long-term time series forecasting. Among the evaluated approaches, the full fine-tuning strategy, which retrains all the parameters, demonstrates superior performance across all tested scenarios with spanning prediction windows.

Furthermore, the integration of the self-supervised pretraining strategy effectively mitigates the error growth rate in extended forecasting horizons. For instance, in the prediction for the subsequent 1-432 steps, the self-supervised learning enhanced SST-iTransformer with full-parameter retraining achieves a significantly lower MSE of 0.396 compared to its pure supervised counterpart (iTransformer-patch) of 0.407, highlighting the advantage of self-supervised representation learning in modeling long-range dependencies.

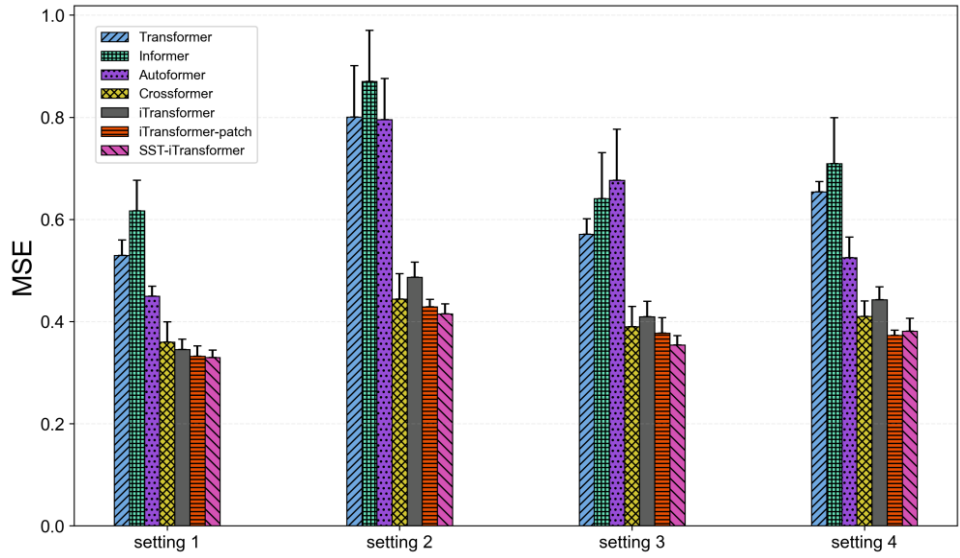

(a)

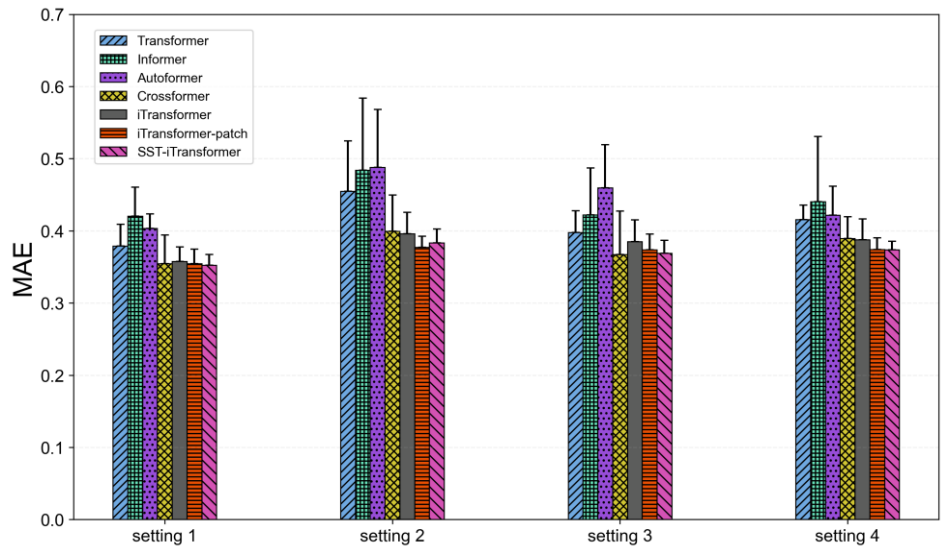

(b)

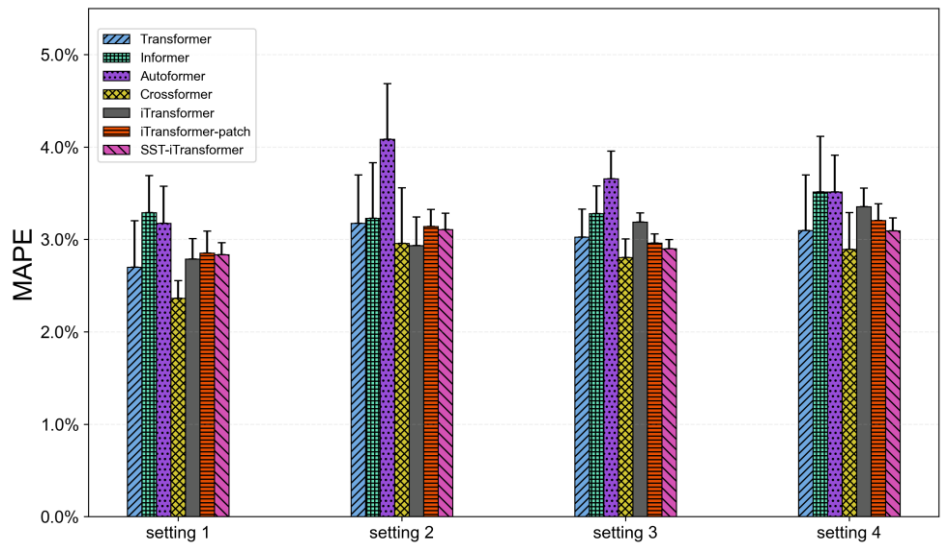

(c)

**Fig. 4**. Performance comparison of evaluated models under varying feature settings through (a) MSE, (b) MAE, and (c) MAPE

**Table 5.** Results for comparative evaluation of fine-tuning strategies across varying prediction steps

| Models | | Predicting steps | | | | | | | | | |
|---|---|---|---|---|---|---|---|---|---|---|---|
| | | 1-144 | | 1-216 | | 1-288 | | 1-360 | | 1-432 | |
| | | MSE | MAE | MSE | MAE | MSE | MAE | MSE | MAE | MSE | MAE |
| Transformer | | 0.529 | 0.379 | 0.542 | 0.384 | 0.573 | 0.396 | 0.576 | 0.400 | 0.702 | 0.448 |
| Informer | | 0.617 | 0.420 | 0.642 | 0.428 | 0.663 | 0.438 | 0.725 | 0.450 | 0.739 | 0.454 |
| Autoformer | | 0.449 | 0.403 | 0.902 | 0.514 | 0.682 | 0.480 | 0.851 | 0.567 | 0.878 | 0.550 |
| DLinear | | 0.592 | 0.419 | 0.622 | 0.430 | 0.651 | 0.437 | 0.670 | 0.441 | 0.697 | 0.449 |
| PatchTST | | 0.364 | **0.349** | 0.372 | 0.372 | 0.390 | 0.379 | 0.407 | 0.384 | 0.416 | <u>0.386</u> |
| Crossformer | | 0.360 | 0.354 | 0.375 | 0.368 | 0.389 | **0.372** | 0.392 | **0.376** | 0.415 | 0.392 |
| iTransformer | | 0.346 | 0.358 | 0.363 | 0.367 | 0.382 | 0.384 | 0.389 | 0.382 | 0.419 | 0.387 |
| iTransformer-patch | | <u>0.332</u> | 0.355 | 0.356 | 0.369 | 0.379 | 0.386 | 0.395 | 0.383 | 0.407 | 0.391 |
| SST-iTransformer | Full fine-tuning | **0.329** | <u>0.352</u> | **0.342** | **0.362** | **0.376** | <u>0.374</u> | 0.393 | <u>0.380</u> | <u>0.396</u> | 0.392 |
| | Linear Probe | 0.421 | 0.409 | 0.437 | 0.419 | 0.461 | 0.429 | 0.481 | 0.435 | 0.500 | 0.444 |
| | Attn-tuning | 0.343 | 0.368 | 0.372 | 0.381 | 0.387 | 0.389 | 0.398 | 0.390 | 0.418 | 0.404 |
| | SAttn-tuning | 0.338 | 0.365 | <u>0.352</u> | 0.377 | 0.379 | 0.384 | **0.380** | 0.387 | **0.393** | 0.400 |
| | CAttn-tuning | 0.341 | 0.356 | 0.364 | <u>0.366</u> | <u>0.377</u> | 0.375 | <u>0.384</u> | 0.381 | 0.412 | **0.385** |

*Note: The best result is shown in **bold**, and the second-best result is <u>underlined</u>.*

*Results represent mean values across the 72 target parking lots among the predicted 1-144 steps, 1-126 steps, 1-288 steps, 1-360 steps, or 1-432 steps.*

In contrast, the Linear Probe fine-tuning approach, which adjusts only the output layer, exhibits substantially inferior performance in long-horizon forecasting compared to other fine-tuning approaches. For example, using Linear Probe for fine-tuning, the MSE of SST-iTransformer rises to 0.500 for the predicting window of 1-432 steps, substantially higher than using other fine-tuning approaches. This discrepancy likely stems from the limited capacity of simple linear probe fine-tuning to capture deep temporal dynamics, revealing the specific requirement of parking forecasting tasks for deeper parameter adaptation rather than a shallow adaptation strategy.

To address the limitations of conventional fine-tuning approaches, this study proposes and tests three innovative attention-preserving fine-tuning mechanisms that strategically freeze specific attention components while fine-tuning remaining parameters. Specifically, this study introduces:

- SAttn-tuning: Freezes the pre-trained sequence attention module while fine-tuning other parameters;
- CAttn-tuning: Freezes the channel attention module while fine-tuning other parameters;
- Attn-tuning: Freezes all attention layers (both sequence and channel) while fine-tuning the remainder.

Experimental results demonstrate the efficacy of these strategies in long-term forecasting. SAttn-tuning achieves the lowest MSE when predicting for the windows of 1-360 and 1-432 steps (0.380 and 0.383, respectively), while CAttn-tuning yields the optimal MAE for the case of 1-432 steps (0.385). Attn-tuning also achieves strong performance in long-term forecasting, maintaining only a small performance gap relative to full fine-tuning and outperforming the Linear Probe approach across extended horizons.

These results robustly confirm that strategically preserving the pretrained parameters in the attention module effectively suppresses error accumulation in extended forecasting horizons. The attention-preserving tuning strategy not only reduces computational demands and accelerates inference for new prediction tasks but also better retains transferable knowledge learned from the original data, thereby enhancing adaptability to parking availability prediction across varying sequence lengths, especially long-term ones, where SAttn-tuning and CAttn-tuning even beat the full fine-tuning.

### 4.3 Sensitivity analysis of temporal context

To evaluate the impact of historical input length (look-back window) on parking availability prediction, this study systematically evaluated model performance across varying look-back windows. Each model was trained to forecast parking availability for the subsequent 1-144 time steps (10 minutes to 24 hours), with comprehensive performance assessment using the aforementioned three complementary evaluation metrics: MSE, MAE, and MAPE. The experimental results, presented in **Fig. 5**, reveal critical insights regarding the relationship between historical context length and prediction accuracy across diverse architectural paradigms.

As illustrated in **Fig. 5**(a) and **Fig. 5**(b), most models demonstrate enhanced prediction accuracy with extended historical sequences, as evidenced by decreasing MSE and MAE values. This pattern aligns with theoretical expectations that longer historical contexts provide richer temporal patterns for model learning. However, certain Transformer variants, specifically Informer and Autoformer, exhibit anomalous performance deterioration with increasing sequence length, contrary to anticipated improvements. This counterintuitive behavior likely stems from excessive global temporal dependency modeling, where extended sequences exacerbate model complexity and introduce prediction instability through overfitting to non-persistent temporal patterns. Notably, Autoformer achieves optimal performance with data within the past 144 time steps as inputs (e.g., a look-back window of 24 hours), which aligns with the fundamental daily periodicity of urban parking dynamics. For Autoformer and similar architectures, extending the input beyond one day disrupts the effective utilization of inherent daily periodic patterns, thereby degrading model performance.

In contrast, models specifically designed for sequential dependency extraction (PatchTST) or channel-wise interaction modeling (Crossformer, iTransformer), together with the proposed enhanced iTransformer-patch and SST-iTransformer, demonstrate more stable performance transitions across varying input lengths. Compared with Informer and Autoformer, these models maintain effective capture of long-term temporal correlations without compromising the utilization of daily periodic patterns, enabling consistent performance improvements with extended input sequences. Notably, the enhanced iTransformer-patch and SST-iTransformer architectures, which incorporate patch-based temporal processing with multi-feature fusion mechanisms, achieve more consistent performance gains across MSE and MAE metrics compared to PatchTST. This observation underscores the effectiveness of the proposed dual-branch attention mechanism and self-supervised pretraining in leveraging extended temporal contexts while maintaining model robustness.

A critical insight arises from the analysis of the relative error metric MAPE. As shown in **Fig. 5**(c), MAPE exhibits a consistent upward trend across most models as the historical context expands. This phenomenon arises because longer input windows enable models to capture large-scale temporal trends more effectively, hence the observed improvements in MSE and MAE, yet may simultaneously reduce accuracy in fitting small-valued cases (e.g., near-zero parking availability). Since MAPE normalizes errors by the true value, even minor deviations in these low-availability periods translate into disproportionately large relative errors. In essence, the models may tend to prioritize broader temporal dynamics (such as daily cycles) at the expense of precision in small-scale fluctuations, thereby increasing MAPE despite improved performance in decreasing absolute error measures (MSE and MAE). This observation underscores the importance of cautious interpretation of relative error metrics and low-availability scenarios in volatile urban parking contexts and highlights the need for further investigation in future studies.

Taken together, these patterns suggest that selecting an appropriate look-back window is critical for balancing prediction robustness and stability. Also, considering the computational complexity and resource constraints, a historical window of 144-288 steps (24-48 hours look-back window) could be a particularly reasonable choice, offering robust and consistent overall performance across models.

Comprehensive comparative analyses confirm that the enhanced iTransformer-patch model achieves measurable performance improvements over the baseline architecture across multiple sequence lengths. More significantly, the SST-iTransformer framework, incorporating self-supervised pretraining and multi-source data fusion through the PCZ, demonstrates superior long-sequence modeling capabilities, maintaining performance stability where other architectures falter. All the above findings validate the proposed architectural design choices and underscore the importance of integrating spatial correlations with temporal modeling for effective parking availability prediction in complex urban environments.

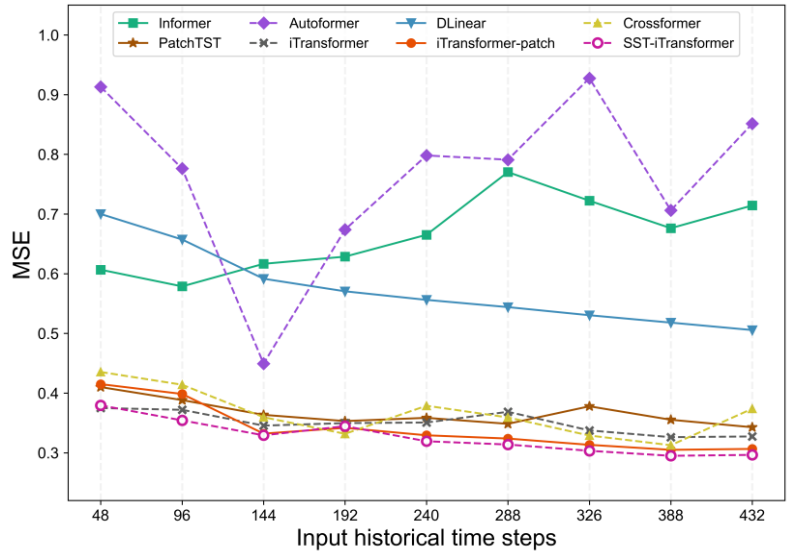

(a)

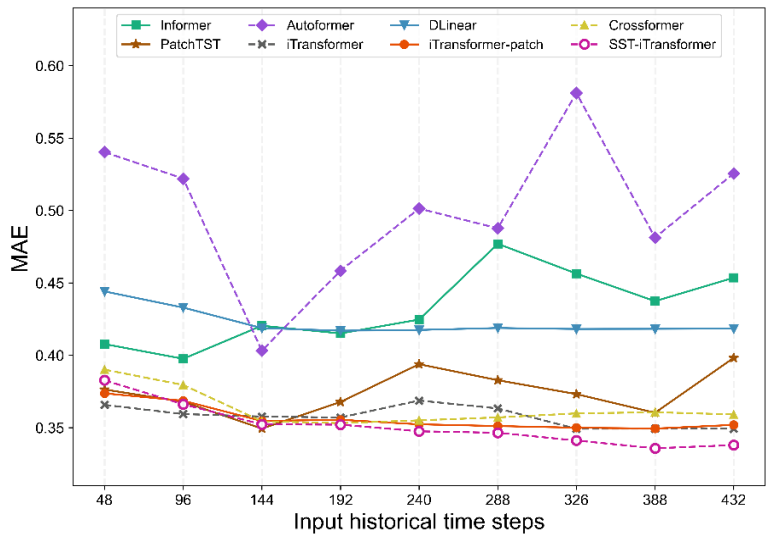

(b)

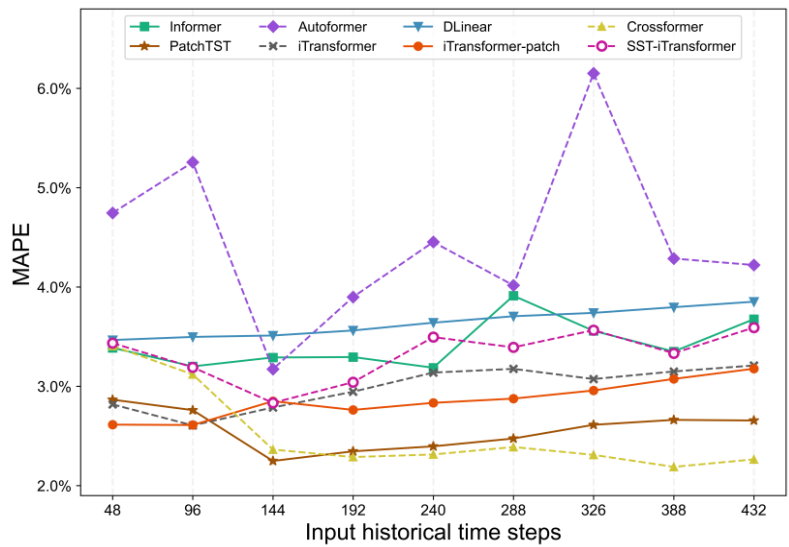

(c)

**Fig. 5.** Performance comparison of evaluated models under varying input lengths through (a) MSE, (b) MAE, and (c) MAPE

# 5. Conclusion

In the era of rapid urbanization, addressing parking challenges is critical to alleviating urban traffic congestion and improving traffic operation efficiency. The proliferation of IoT, ICT, and Artificial Intelligence technologies provides unprecedented opportunities for developing accurate and reliable data-driven parking availability prediction systems. While prior ML and DL approaches have demonstrated potential, their effectiveness remains constrained by suboptimal feature engineering, inadequate integration of multimodal transportation synergies, and insufficient modeling of cross-lot dependencies within volatile spatio-temporal contexts.

To bridge these critical gaps, this study proposes a Self-supervised learning enhanced Spatio-temporal Inverted Transformer (SST-iTransformer) framework that systematically captures spatial interdependencies among relevant proximate parking lots while modeling dynamic temporal patterns through a novel dual-attention architecture (i.e., Series Attention for long-range dependencies and Channel Attention for cross-variate interactions). Parking cluster zones (PCZs), derived via K-means clustering, are employed to group related lots and integrate multi-source mobility demand features from, e.g., metro, bus, ride-hailing, taxi, and parking data. Through rigorous experimentation using real-world data from Chengdu, China, the proposed methodology demonstrates significant advancements over existing approaches.

Extensive experiments benchmarked the proposed SST-iTransformer against baselines of RNNs (GRU, LSTM), Transformer variants (Informer, Autoformer), channel-independent models (DLinear, PatchTST), and channel-dependent models (Crossformer, iTransformer). Results demonstrated SST-iTransformer's superior performance, achieving the lowest MSE, alongside competitive MAE across forecasting horizons. Rigorous ablation studies quantitatively established the relative importance of different supplement data sources: ride-hailing demand features demonstrated the most influential (contributing to the largest performance degradation when omitted), followed by taxi demand, whereas fixed-route transit feature data provided a marginal impact for the parking prediction task. Moreover, modeling spatial dependencies within PCZs proved indispensable; omitting correlated parking lot histories led to severe performance degradation, underscoring the critical importance of the proposed clustering approach.

Furthermore, in long-term parking series forecasting, both sequence-specific models and channel-enhanced architectures, as well as the proposed SST-iTransformer, are prone to progressive error accumulation, which substantially degrades prediction accuracy. To mitigate this issue, three attention-preserving fine-tuning strategies were introduced. By selectively freezing pre-trained attention modules while refining the remaining parameters, these strategies suppress error propagation over extended horizons, preserve representational capacity, and enable rapid adaptation to new operational scenarios. Importantly, they also reduce computational overhead, making the approach viable for real-world deployment in city-scale intelligent parking systems.

Collectively, these findings provide critical insights for high-volatility parking prediction, demonstrating that the synergistic integration of multi-source data fusion with self-supervised learning significantly enhances forecasting robustness. The proposed framework not only advances the theoretical understanding of spatio-temporal dependencies in parking dynamics but also offers practical utility for urban planners and transportation authorities. By enabling more accurate parking availability prediction, this research facilitates the development of intelligent parking management systems that can optimize resource utilization, reduce congestion, and promote sustainable urban mobility. Future research directions include extending the framework to incorporate real-time event data and exploring its application in dynamic pricing mechanisms for parking resources.

## CRediT authorship contribution statement

**Yin Huang**: Conceptualization, Data curation, Formal analysis, Investigation, Methodology, Validation, Visualization, Writing – original draft, Writing – review & editing. **Yongqi Dong**: Conceptualization, Formal analysis, Investigation, Methodology, Supervision, Validation, Visualization, Writing – original draft, Writing – review & editing. **Youhua Tang**: Conceptualization, Data curation, Funding acquisition, Project administration, Resources, Supervision, Validation, Writing – review & editing. **Li Li**: Funding acquisition, Project administration, Resources, Validation, Writing – review & editing.

**Declaration of competing interests**

The authors declare that they have no known competing financial interests or personal relationships that could have appeared to influence the work reported in this paper.

**Acknowledgments**

This work was supported by the Natural Science Foundation of Beijing, China, Grant Number: L231007.